%% file: root.tex
\documentclass[runningheads]{llncs}
\usepackage[T1]{fontenc}
\usepackage{subcaption}
\usepackage{amsmath}
\usepackage{amsfonts}
\usepackage[bookmarks=true]{hyperref}
\usepackage{graphicx}
\usepackage{times}
\usepackage{underscore}
\usepackage{moreverb,url}
\usepackage[dvipsnames]{xcolor}
\usepackage{url}
\usepackage{times}
\usepackage{multicol}
\usepackage[utf8]{inputenc}
\usepackage{graphics}
\usepackage{bm}
\usepackage[linesnumbered,ruled,noend]{algorithm2e}
\usepackage[noend]{algpseudocode}
\usepackage{multicol}
\usepackage{multirow, makecell}
\usepackage{lipsum}
\usepackage{dblfloatfix}
\usepackage[english]{babel}
\usepackage{blindtext}
\usepackage{mathtools}
\usepackage{booktabs}
\usepackage{subcaption}
\usepackage{enumerate}
\usepackage{thmtools}
\usepackage{cleveref}

\SetKwRepeat{Do}{do}{while}
\DeclareMathOperator*{\argmax}{arg\,max}
\usepackage{multirow}
\usepackage{booktabs}
\usepackage{siunitx}
\usepackage{caption}

\usepackage{float} % To fix the error: "LaTeX Error: Unknown float option `H'."

\newcommand{\segm}{\cA}
\newcommand{\pick}{\text{pick}}
\newcommand{\place}{\text{place}}
\newcommand{\grasp}{\text{grasp}}
\usepackage[backend=biber,
            hyperref=true,
            url=false,
            isbn=false,
            doi=false,
            backref=false,
            style=ieee,
            natbib=true,%compatibility aliases
            mincitenames=1,
            maxcitenames=1,
            citestyle=numeric-comp,
            sorting=nyt,%none
            block=none,
            maxbibnames=99]{biblatex}
\usepackage{hyperref}

\usepackage{afterpage}

\usepackage{resources/custom_commands}

\usepackage{tikz}
\newcommand*\circled[1]{\tikz[baseline=(char.base)]{
            \node[shape=circle,draw=white,inner sep=2pt,line width=0.3 mm, text=white] (char) {#1};}}
            
\usepackage{booktabs, siunitx}

\newcommand{\pred}{p}
\newcommand{\revision}[1]{\textcolor{black}{#1}}
\newcommand{\annotate}[1]{\textcolor{black}{#1}}

\begin{document}

\title{The Teenager’s Problem:  Efficient Garment Decluttering 
as Probabilistic Set Cover
% with Grasp Optimization
}
\titlerunning{The Teenager’s Problem}
%\author{First Author\inst{1}\orcidID{0000-1111-2222-3333} \and
%Second Author\inst{2,3}\orcidID{1111-2222-3333-4444} \and
%Third Author\inst{3}\orcidID{2222--3333-4444-5555}}

\author{Aviv Adler\thanks{Equal Contribution}\inst{1} \and Ayah Ahmad$^{\star}$\inst{1} \and Yulei Qiu$^{\star}$\inst{2} \and Shengyin Wang\inst{2} \and Wisdom C. Agboh\inst{1}$^,$\inst{2} \and Edith Llontop\inst{1} \and Tianshuang Qiu\inst{1} \and
Jeffrey Ichnowski\inst{3} \and 
 Thomas Kollar\inst{4} \and Richard Cheng\inst{4} \and Mehmet Dogar\inst{2} \and Ken Goldberg\inst{1}
}

\authorrunning{A. Adler et al.}
\institute{University of California, Berkeley, USA \and
University of Leeds, UK \and
Carnegie Mellon University, USA \and
Toyota Research Institute, USA
}
\maketitle              % typeset the header of the contribution
\begin{abstract}
This paper addresses the ``Teenager's Problem'': efficiently removing scattered garments from a planar surface into a basket. As grasping and transporting individual garments is highly inefficient, we propose policies to select grasp locations for multiple garments using an overhead camera. 
Our core approach is \emph{segment-based}, which uses segmentation on the overhead RGB image of the scene. We propose a \emph{Probabilistic Set Cover} formulation of the problem, aiming to minimize the number of grasps that clear all garments off the surface. Grasp efficiency is measured by \emph{Objects per Transport} (OpT), which denotes the average number of objects removed per trip to the laundry basket. 
Additionally, we explore several \emph{depth-based} methods, which use overhead depth data to find efficient grasps. Experiments suggest that our segment-based method increases OpT by $50\%$ over a random baseline, whereas combined \emph{hybrid} methods yield improvements of $33\%$. Finally, a method employing \emph{consolidation} (with segmentation) is considered, which locally moves the garments on the work surface to increase OpT, when the distance to the basket is much greater than the local motion distances. This yields an improvement of $81\%$ over the baseline.

\end{abstract}

\input{sections/01-intro-motivation-v2}

\input{sections/02-related-work}

\input{sections/03-problem-statement}

\input{sections/04-methods}

\input{sections/05-NNs}

\input{sections/06-Results}

\input{sections/07-dirty-laundry-conclusion}

\printbibliography

\end{document}

%% file: sections/01-intro-motivation-v2.tex
\section{Introduction}

We introduce the ``Teenager's Problem'': removing a large number of scattered garments from a surface (e.g. the floor of a teenager's room, or a work surface) in the shortest time. This problem has applications in hotels, retail dressing rooms, garment manufacturing, and other domains where heaps of garments must be manipulated efficiently. This general problem can also be applied to cases where other deformable objects must be removed, for instance in clearing litter or garden debris.

\begin{figure}[t!]
    \centering
    \includegraphics[width=0.5\linewidth]{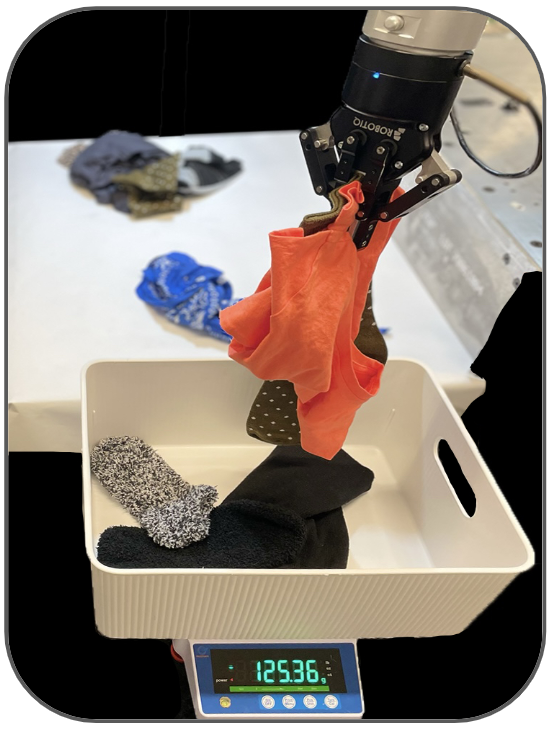}
    \caption{An instance of the \textit{Teenager's Problem} in the experimental setup; the work surface is white and the basket is beige, a UR5 industrial robot with a Robotiq parallel-jaw gripper is used, with overhead cameras above. The scale automatically records weight data as experiments are run. }
    \label{fig:splash} 
    %\vspace{-2pc}
\end{figure}

We first formalize the Teenager's Problem, then consider several methods to solve it. Consider Fig.~\ref{fig:splash} with multiple garments on a work surface. Given an overhead RGB or RGBD image, what robot pick-and-place actions would minimize the total time to move all of the garments to a laundry basket? Removing garments one-by-one would be inefficient. Thus we propose that the robot should use the deformable nature of garments and grasp multiple garments at once. 

Given a scene like the one in Fig.~\ref{fig:splash}, one approach is to identify individual garments and optimize grasps to pick as many of these garments as possible. This motivates \textit{segment-based} methods, i.e., methods that use the RGB image to segment the individual garments. We use the Segment Anything Model (SAM) \cite{segment-anything} to approximately model individual garments. Then, given the set of garments (segments) and a candidate grasp point, our method predicts the probabilities that each garment will be picked by that grasp. For example, a grasp candidate that is close to the intersection of a set of garments would have a positive probability of picking each of those garments, while it would have near-zero probability of picking garments that are farther away. Given many such candidate grasps, and their predicted probabilities of picking each garment, we formulate the task of minimizing the number of grasps to pick all garments as a \textit{Probabilistic Set Cover Problem} \cite{ahmed2013probabilistic}. We then express this as a Mixed Integer Linear Program and solve it.

While the segment-based method is able to identify grasps that would pick multiple \textit{visible} garments simultaneously, it can also miss some multi-garment grasps. In particular, since only the top surface of the garments is visible to the camera, garments that are fully occluded are ignored by the segment-based method.

A second approach to solving the Teenager’s Problem is to  treat the whole scene as a homogeneous volume to be removed. This motivates \textit{depth-based} methods, i.e., methods that use the depth image to infer grasp points that would then remove as much volume as possible. We consider two depth-based methods in this paper. The first method uses \textit{height} and grasps at the highest garment point in the scene. The second method estimates a \textit{volume}, by integrating the depth data within a grasp radius, and grasps at the point in the scene that gives the largest estimated volume. 

The depth-based methods can identify large heaps and grasp multiple garments at a time, even when some of these garments are completely occluded by others and not individually visible to the camera. However, the depth-based methods also miss some good grasps. Particularly, since they do not detect individual garment positions and boundaries, they miss grasp points that  may pick multiple garments simultaneously but are at a lower height or volume, e.g., points where multiple garment boundaries meet.

We also consider a \textit{hybrid} method to combine the complementary strengths of the segment- and depth-based approaches. The hybrid method chooses which %uses either a depth-based method or a segment-based
method
depending on the maximum height available.

Finally, we consider methods that make use of \emph{consolidation} actions that move within the workspace to gather the garments into larger heaps, before removing the heap. This improves the efficiency of grasps to transport the objects to the basket at the cost of the time used in consolidation, which can improve efficiency in cases where the basket is located far from the work surface. 

Experiments suggest that, the segment-based method significantly reduces robot trips to the basket by $50\%$. The \emph{hybrid} methods yield improvements of $33\%$. Finally, Objects per Transport (OpT) can be further increased by $81\%$ by using \emph{consolidation} actions within the workspace to set up highly efficient transport actions.

We make the following contributions:
\begin{itemize}
    \item A formalization of the Teenager's Problem as a Probabilistic Set Cover Problem.
    \item Five methods (two depth-based, one segment-based, and two hybrid) to generate effective multi-garment grasps.
    \item A method that uses heap consolidation along with the segment-based grasp generation method to efficiently solve the Teenager’s Problem.
    \item Physical experiments and data from grasping 1750 garments, that compare the performance of the six methods against a random baseline. 
\end{itemize}

%% file: sections/02-related-work.tex
\section{Related work} 
Our work is related to two lines of work: \textit{manipulation of deformable objects} and \textit{multi-object manipulation}.

\subsection{Manipulation of Deformable Objects}

Prior work on deformable object manipulation includes folding \cite{Yahav-IROS-2022,Hoque2022},
flinging \cite{zhang2021robots,ha2022flingbot,chen2022efficiently}, 
fabric smoothing \cite{Seita-IROS-2020,9561980,Sharma2022}, bed-making \cite{Seita2018DeepTL}, untangling ropes \cite{SundaresanGrannen-RSS-21}, and singulating clothes from a heap \cite{Willimon-IEEE-2011,Sashank-IROS-2022}. Several works aimed to detect specific features, such as the corners and edges of fabrics, and to identify optimal grasp points \cite{9341121,Maitin-ICRA-2010,9945180}. Other techniques employ deep learning to identify successful grasps \cite{Lenz2013,Chu2018}. Some studies have focused on determining optimal grasp points by considering not only the depth of the cloth but also targeting wrinkles as highly graspable regions \cite{Ramisa2011DeterminingWT,Ramisa-IEEE-2012,Wang-IEEEAccess-2020,qiu2023robotic}. These prior works focus on manipulating a single deformable object at a time or singulating a deformable object from among others. Our work, on the other hand, concentrates on grasping multiple garments simultaneously. \revision{There also exist specialized grippers developed for garment grasping \cite{le2013development,dragusanu2022dressgripper}. While the strategies we propose can also be used with such specialized grippers, we experiment here with a general-purpose gripper,} \annotate{which increases the applicability of the results}.

\subsection{Multi-object Manipulation}

 Multi-object grasping can improve decluttering efficiency \cite{Agboh-ISRR-2022}. It has been studied, with analytic methods \cite{Yamada-JCSE-2015,yao2023exploiting}, learning-based methods \cite{Agboh-arxiv-2022, Chen2021}, and with special gripper designs \cite{Nguyen-AME-2022}. The focus, however, has remained on rigid objects. Instead, we consider the problem of grasping multiple deformable objects at a time, and using such grasps to efficiently clear a surface. 
 
Multi-object manipulation scenarios can encompass cluttered \cite{Kasaei2021} environments, which can include both deformable and rigid objects \cite{9139227}, however, their goal is to singulate the objects to grasp them individually.
% Grasping multiple objects can occur in cluttered or decluttered spaces \cite{Tirumala-IROS-2022}\cite{Kasaei2021} , and can contain a mix of deformable and non-deformable objects \cite{9139227}. 
Prior work on manipulating multiple rigid objects used methods such as pushing, stacking, and destacking \cite{Huang-ISRR-2022,Agboh-ISRR-2022,Sakamoto-IROS-2021}.
% These methods have served as a motivation to manipulate deformable objects. While these methods have been applied to deformable objects, the problems they sought to solve were different--such as identifying objects in a cluttered space 
%While these techniques served as a foundation for addressing deformable objects, their initial focus lay in different problem domains, such as object identification in cluttered spaces \cite{Willimon-IEEE-2011}. 
In cluttered scenes with multiple rigid objects, one method for determining how, or where, to grasp is by using image segmentation \cite{6088647}, detecting and isolating individual objects in the scene. We also use a segmentation approach but for deformable objects.

%% file: sections/03-problem-statement.tex
\section{The Teenager's Problem} \label{sec:problem-statement}

We formulate the \textit{Teenager's Problem} as follows: deformable objects rest on a planar work surface. Given a fixed target basket, the goal is to transfer all the garments efficiently from the workspace to the basket with a minimum number of grasps (which naturally maximizes OpT). %This means executing a series of pick-and-place motions, which fall into two categories: first, movements from the workspace to the target basket ("transports") and second, optional movements within the workspace ("consolidation") of the garments to increase future OpT.

\subsection{Problem Statement}

In the Teenager's Problem, there are $m$ differently colored garments on a surface. We assume we are given $n$ grasp candidates, where each grasp is a tuple $(x,y,\theta) \in \bbR^2 \times [-\pi/2, \pi/2]$ representing a top-down grasp, with $\theta$ representing the angle of the parallel-jaw gripper with respect to the workspace axes. (We explain how we generate such grasp candidates in Sec.~\ref{sec:segment-based}.) We denote a \textit{grasp plan} as $\bx \in \{0,1\}^n$, with $x_i$ indicating whether grasp candidate $i$ is in the grasp plan; i.e. $x_i = 1$ if it is in the plan and $x_i = 0$ otherwise. 

For a grasp $i$ and garment $j$, we denote by $p_{i,j}$ the probability that grasp $i$ successfully grasps garment $j$. (We describe how we estimate $p_{i,j}$ in Sec~\ref{sec:segment-based}.)
For each garment $j$, we denote by $q_j$ the minimum probability that we want to have of removing garment $j$ from the surface.

Then, the objective is to find a grasp plan $\bx$ that minimizes $\bone^\ltop \bx$ (i.e. the number of grasps in the plan), such that for all garments $j \in \{1, 2, \dots, m\}$, thus
\begin{align}\label{eq:p_remove_garment}
    \bbP[\text{remove garment } j \, | \, \bx] \geq q_j. % Note: the period is intentional, punctutate equations when they are part of sentense.
\end{align}

The Teenager's Problem can thus be seen as a probabilistic variant of the classic Set Cover problem, since each grasp corresponds to the (weighted) set of garments it could potentially grasp, and the goal is to find a set of grasps whose `union' encompasses all the garments. This also means that the benefit of a grasp depends on the set of other grasps that will also be taken---even if a grasp is likely to get several garments, it may be useless if those garments were already likely removed by other grasps in the set.

\subsection{Mixed Integer Linear Program}\label{sec:milp}

We propose to solve the above problem by converting it to a Mixed Integer Linear Program (MILP).

Eq.~\ref{eq:p_remove_garment} is equivalent to
\begin{align}
    \bbP[\text{fail to remove garment } j \, | \, \bx] \leq 1-q_j. % Note: the period is intentional, punctutate equations when they are part of sentense.
\end{align}

For any two grasps $i, i'$ that do \textit{not} overlap (i.e., grasps that are sufficiently away from each other), we assume the success probabilities are independent random events. Therefore, in any grasp plan without overlapping grasps, the success or failure of any grasp in getting any garment can be treated as independent from any other. We call grasp plan without overlapping grasps \emph{valid}. {\color{black} We will restrict the planner to valid plans in order to prevent the planner from choosing grasps which are too close together and would likely interfere with each other.}

Now we consider a valid grasp plan $\bx = (x_1, \dots, x_n) \in \{0,1\}^n$. Then the probability of failing to get garment $j$ is
\begin{align}
    \bbP[\text{fail to remove garment } j \, | \, \bx] = \bbP[\text{no } i \text{ s.t. } x_i = 1 \text{ picks } j] \\
    = \prod_{i : x_i = 1} (1 - p_{i,j}) 
    = \prod_i (1 - p_{i,j})^{x_i}. % Note: the period is intentional, punctutate equations when they are part of sentense.
\end{align}
Therefore we want $\bx$ satisfying the constraints such that
\begin{align}
    \prod_i (1 - p_{i,j})^{x_i} &\leq 1-q_j
    \\ \iff \sum_i x_i \log(1 - p_{i,j}) &\leq \log(1-q_j). % Note: the period is intentional, punctutate equations when they are part of sentense.
\end{align}
We can assume that $p_{i,j} < 1$ since no grasp is 100\% certain of success, and $q_j < 1$ since we cannot get any garment with 100\% certainty either, hence we can assume all the values above are finite.

Thus, we define an MILP using matrix $\bA$ where $A_{i,j} = \log(1-p_{i,j})$ and $b_j = \log(1-q_j)$. We want to minimize the number of grasps used, %so the objective function vector will be $c_i = 1$ for all $i$. This 
which gives the MILP:
\begin{equation}\label{eq:milp}
    \begin{aligned}
        \text{minimize } \bone^\ltop \bx & \text{ subject to}
        \\ \bA^\ltop \bx &\leq \bb
        \\ x_i + x_{i'} &\leq 1 \text{ for all } i, i' \text{ which overlap}
        \\ x_i &\in \{0,1\} \text{ for all } i \,, % Note: the comma is intentional, punctutate equations when they are part of sentense.
    \end{aligned}
\end{equation}
where the second constraint enforces that the grasp plans considered are valid.

In our implementation, we set $q_j=0.7$ as the minimum probability of a garment being removed. We use the MILP solver \cite{huangfu2018parallelizing} within the SciPy library \cite{2020SciPy-NMeth}.

\subsection{Metrics}

The primary metric for evaluating the methods is \emph{Objects per Transport} (OpT), which denotes the average number of objects taken during each transport and measures the general effectiveness of the performed grasps. We use OpT, as opposed to Picks Per Hour (PPH), as OpT directly measures grasp quality and PPH depends heavily on implementation details (particularly concerning computation time) which reduces its reliability as a metric in this setting.

%% file: sections/04-methods.tex
\section{Teenager's Problem Methods}

This study explores various strategies for efficiently grasping multiple garments concurrently, categorized broadly into two types:  \emph{segment-based} and \emph{depth-based} approaches. 

All the methods described below use a pre-processing of the RGB pixels to separate the background and the foreground, i.e. to determine the \emph{garment points}, denoted $\cX_g$. Since we assume the system knows the color and/or pattern of the background, this is achieved with color thresholding.

\subsection{Segment-Based Methods} \label{sec:segment-based}

\begin{figure*}[!ht] \centering \includegraphics[scale=0.42
]{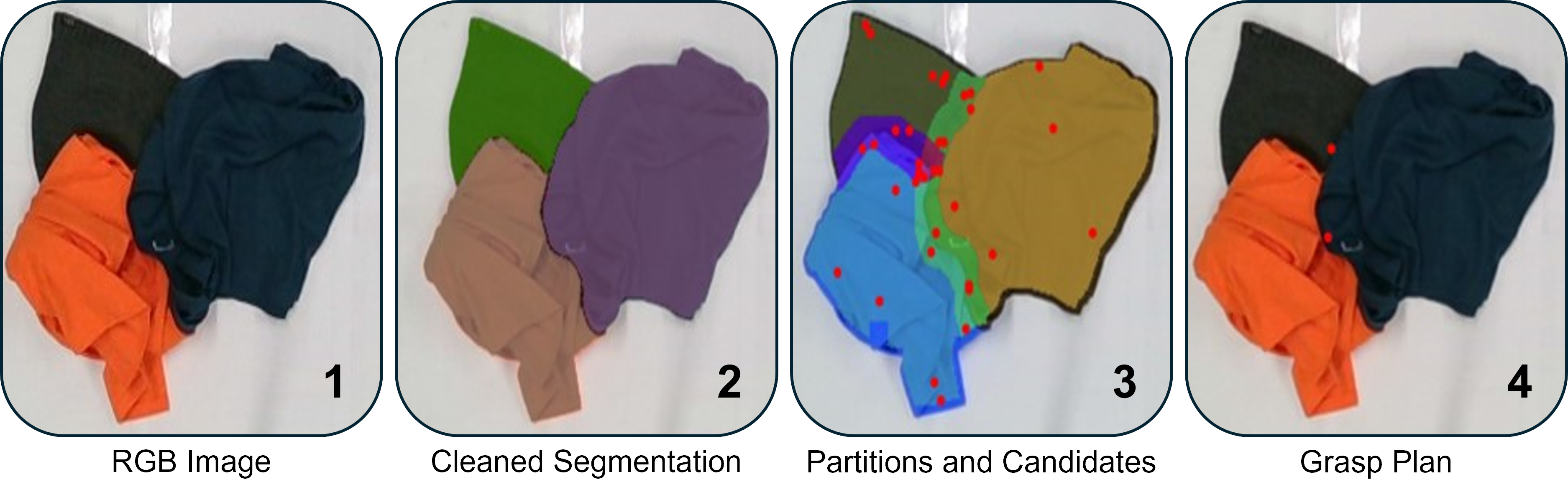} \caption{An example of the segment-based grasp point selection algorithm (grasp orientations not shown).
%(Alg.~\ref{alg:segmentation-declutterer}), 
 %through selection of grasp points (step (2))
 %, where Alg.~\ref{alg:candidate-pt-selection} is run),
 %before orientations are selected.
 {\bf From left to right:} \emph{(1)} The original overhead RGB image. \emph{(2)} The cleaned segmentation $\cM$. \emph{(3)} The partitions such that, each partition has the same `nearby' set of segments; along with $k=5$ grasp candidates sampled from each partition (red dots). \emph{(4)} The two grasps (red dots) that constitute the grasp plan, found by solving the MILP.} \label{fig:seg-algo-seq} \end{figure*}

The segment-based approach divides the task into \emph{cycles}: a segmentation is generated at the start of each cycle, along with candidate grasps. Using this segmentation and the grasp candidates, the MILP  (Eq.~\ref{eq:milp}) is solved, and the sequence of grasps is carried out based on that initial segmentation (however, an overhead RGB image is still taken between each grasp, as will be explained later). Each cycle relies on four subroutines: 
(1) {Segmentation and cleanup}; 
%where an RGB image of the workspace is divided into segments, ideally representing individual garments; 
(2) {prediction}; 
%which predicts what segments will be removed given a grasp point and orientation; 
(3) {grasp selection}; 
%which chooses grasps based on the segmentation and predictor with the goal of removing as many segments as possible with few grasps; 
(4) execution. Once all the planned moves have been performed the next cycle begins.
This cycle is repeated until the table is cleared. %One important point is that while a new RGB image is taken after every basket move, a new segmentation is only computed at the start of each cycle; this is because many of the grasps planned during a cycle are still effective even after other grasps have been performed.
%which performs the grasps from the closest to the basket to the furthest, skipping grasps whose `neighborhoods' (nearby areas) have been affected by previous grasps.
The steps of a cycle are detailed below.

\subsubsection{Segmentation and cleanup}

The method uses Meta's Segment Anything Model (SAM) \cite{segment-anything} with the \emph{vit-b} weights, prompts the image as a whole, and generates segmentation masks for the image. However, the initial segmentation often contains multiple overlapping segments, gaps, and regions corresponding to the work surface. To address this, our method begins by applying color thresholding to remove the segments that represent the work surface. The remaining clothes and background regions are then converted into a binary cloth map. To effectively identify and fill/remove holes within the cloth map, we convolve this map twice using an all-ones kernel. 

% However, the returned segmentation generally contains multiple overlapping segments (for instance, both a whole glove and its individual fingers might all be segments) and spaces with no segment, and regions corresponding to the work surface itself (as opposed to garments) are usually given segments. After generating the initial segmentation, the method removes redundant segments (in order of size from largest to smallest) and fills in large contiguous gaps as their own segments, and finally uses color thresholding to remove segments corresponding to the work surface.

Let $\cM$ denote the set of segments detected using this method. For that cycle, the number of garments is set to be the number of detected segments, i.e., $m=|\cM|$, where each $M_j \in \cM$ represents the segment belonging to garment $j$. Fig.~\ref{fig:seg-algo-seq}-(2) shows an example output of this step.

\subsubsection{Prediction}

The Teenager's Problem keeps the predictor function $\pred$ general to accommodate a variety of different approaches, both analytic and learned, to estimating the effects of a grasp. In this work we use an analytic predictor which models the area under the gripper as an ellipse and the probability of a successful grasp as depending on the total area of the garment within the ellipse. 

Specifically, given a grasp $(x,y, \theta)$, let $E(x,y,\theta)$ denote the ellipse centered at $(x,y)$ whose major axis is oriented at angle $\theta$ with major axis length $d_1$ and minor axis length $d_2$. The axis lengths $d_1$ and $d_2$ are scaled to be the length and width of the parallel-jaw gripper. We estimate the probability of successfully grasping a garment $j$ with grasp $i$, as
\begin{align*}
    \pred_{i,j} = \frac{\text{area}(E(x,y,\theta) \cap M_j)}{\text{area}(E(x,y,\theta) \cap M_j) + b} 
\end{align*}
where $(x,y, \theta)$ represent the grasp $i$ and $b > 0$ is a normalization constant. In our implementation, the area is measured in pixels and we use $b=100$.

This predictor is intended to capture the following intuition: the gripper directly affects the area under it, and the more any segment falls in that area, the more likely it is to be removed by the grasp (but cannot have probability $> 1$ of being removed). Note that this predictor can never be 100\% certain that a given segment will be removed, which is realistic.

One property of this predictor is that, while it is important to choose grasps which get a large total area of the segments within the ellipse, having several different segments with nonzero area under the ellipse is generally preferable to having just one (even if the total area within the ellipse is the same), because the benefit of added area for one segment decreases as the area already captured increases. Thus, the best grasps will generally occur at or near the boundary between multiple segments.

\subsubsection{Grasp selection}
% With a clean segmentation and predictor, the method faces the task of determining which grasps to execute from an infinite number of potential options. To simplify the problem, the method selects a set of candidate points (excluding orientations). These candidate points should be individually effective and diverse, ensuring that grasps can capture different segments. An ideal candidate point is situated close to various segments, considering the gripper's radius ($4.25$ cm or 20 pixels). Diverse candidate points encompass distinct segments.
% Given a clean segmentation and a predictor, the question remains: What grasp(s) should the method execute? Even with the ability to predict the effect of any given grasp on the scene, some selection method is needed since there are infinitely many potential grasps. 
With a clean segmentation and predictor, the method faces the task of determining which grasps to execute. 
We first generate a set of candidate grasp points and then use these candidates to solve the MILP (Eq.~\ref{eq:milp}), to generate a grasp plan, $\bx$.

The candidate grasp point generation method 
%given in Alg.~\ref{alg:candidate-pt-selection}, 
does the following (given a radius $r > 0$ in pixels, and segmentation $\cM$):
\begin{itemize}
    \item For each pixel $(x,y) \in \cX_g$, determine the set $S(x,y) \subseteq \cM$ of segments which are within distance $r$ from $(x,y)$.
    \item Partition $\cX_g$ based on $S(x,y)$. In other words, two points $(x,y) \in \cX_g$ and $(x',y')\in \cX_g$ are considered in the same partition, if $S(x,y) = S(x',y')$.
    \item From each partition, randomly (uniform) sample $k$ grasp points $\{(x_i, y_i)\}_{i=1}^k$. We used $k=5$.
    %\item Construct the set $\cS$ of \emph{maximal sets} $S(x,y)$, i.e. $\cS$ consists of every $S(x,y)$ which is not a proper subset of some other $S(x',y')$.
    %\item For each maximal set $S_i \in \cS$, choose a (uniformly) random $(x_i,y_i)$ such that $S(x_i,y_i) = S_i$.
    %\item Return $\{(x_i, y_i)\}_{i=1}^k$, where $k$ is the number of maximal sets.
\end{itemize}
The rationale behind this procedure is to sample candidate grasp points from a variety of regions (partitions) with potential to grasp different garments. While a completely random sampling of $\cX_g$ can miss important regions that have the potential to grasp multiple garments at a time, partitioning  first, and then randomly sampling these partitions helps maintain a rich variety of candidate grasps. Fig.~\ref{fig:seg-algo-seq}-(3) presents an example, where the partitions as well as the sampled grasp candidates are shown. 

Then, for each point $(x,y)$, we generate $\ell$ grasp candidates $(x,y,\theta)$, by enumerating a list of $\ell$ equally-spaced orientations in $[-\pi/2,\pi/2]$. For a balance between efficiency and thoroughness, we used $\ell = 6$.

%Then the procedure chooses an orientation for each grasp point using a greedy heuristic. First, it enumerates a list of $\ell$ equally-spaced orientations in $[-\pi/2,\pi/2]$ and runs the predictor for each orientation on all the segments; the orientation which is predicted to remove the largest number of masks (i.e. the sum of the predicted probability of removal over all the masks) is then chosen for each grasp. For a balance between efficiency and thoroughness, we used $\ell = 6$.

%The ellipses, $E(x,y,\theta)$, of two grasp are used to decide whether the two grasps \textit{overlap}, as discussed in Sec.~\ref{sec:milp}.

Furthermore, if two grasps belong to the same partition, we treat them as \textit{overlapping},  as discussed in Sec.~\ref{sec:milp} (second constraint of Eq.~\ref{eq:milp}).

Fig.~\ref{fig:seg-algo-seq}-(4) shows an example grasp plan (including only two grasps) as generated by the MILP solver. 

\subsubsection{Grasp execution}

The algorithm then attempts all grasps in the grasp plan $\bx$ in sequence, in increasing order of distance to the basket; this is to prevent, as far as possible, dragging garments from disturbing the positions of the garments that remain (which may cause garments to fall off the work surface).

An RGB image is also captured after each transfer to the basket (when the arm is out of frame), although a new segmentation is \emph{not} generated (until the next cycle). Instead, for each planned grasp remaining, the difference between its current state and the state at the beginning of the cycle (when the segmentation was generated) is estimated using the squared difference between the pixel values within a small square neighborhood around the grasp point; if the difference is too large, the grasp is deemed to be in a different state from when it was planned and is not performed. This ensures that grasps are not performed unless the system knows that the local configuration of the garments is approximately the same as when it was planned.

\begin{figure}[t!]
    \centering
    \includegraphics[width=\linewidth]{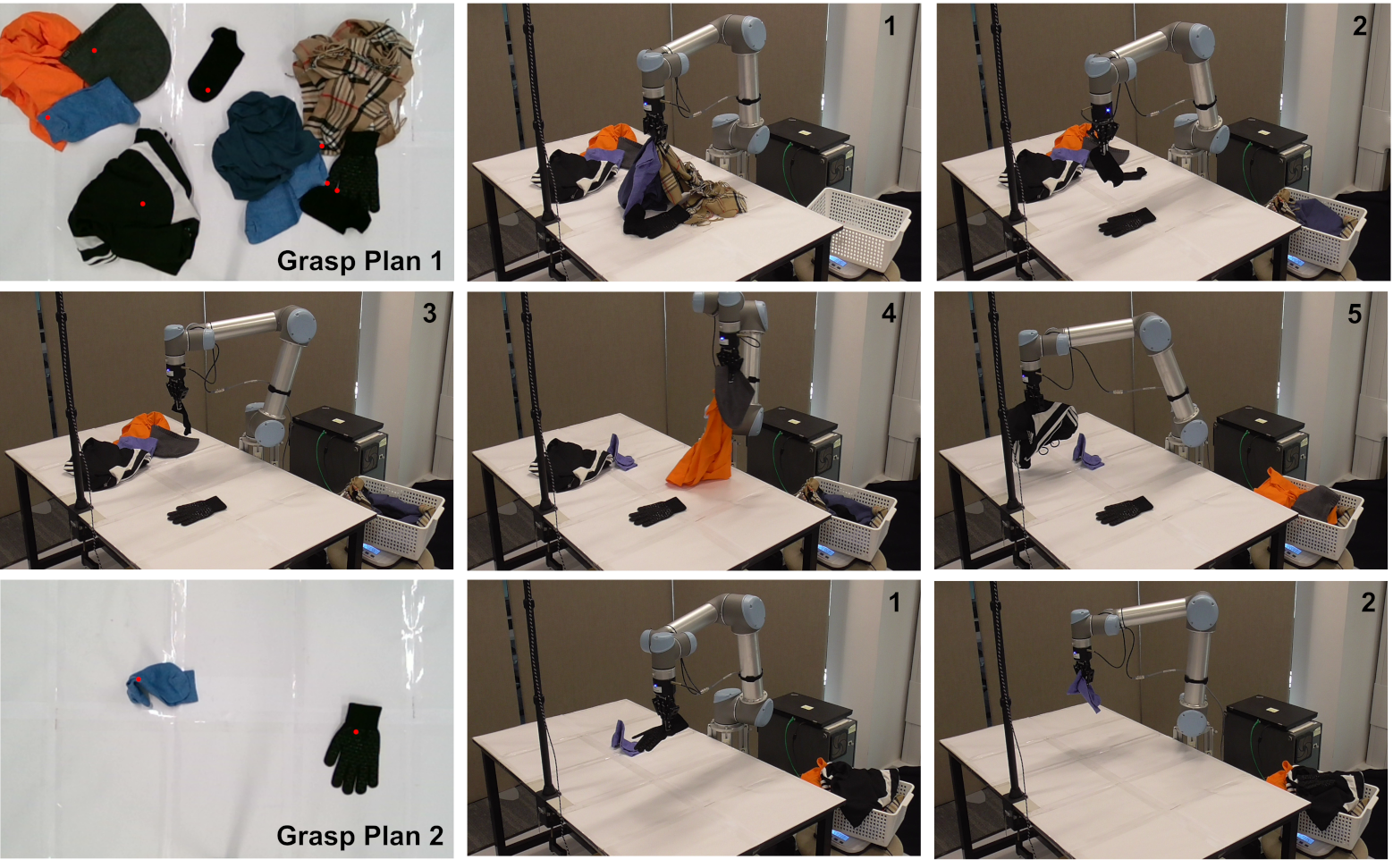}
    \caption{An example execution using the segment-based method, where the table is cleared in seven grasps in total. Grasp Plan 1 is generated on the initial scene (red dots show planned grasp positions). Only five of these grasps are executed, before Grasp Plan 2 is generated with the remaining two grasps to clean the table.} 
    \label{fig:result-seg} 
\end{figure}

This process is visualized in Fig.~\ref{fig:result-seg}, where the system needs two cycles, and seven grasps, before clearing the table. Notice how the system starts from the grasp that is closest to the basket. However, since the first grasp changes the scene significantly in its neighborhood, the system does not execute the nearby two grasps. Similarly, after the grasp of the orange shirt and the hat (Grasp 5 in the figure), the grasp on the blue sock is not attempted. Instead, a new cycle begins requiring only two grasps.

%These points should be individually effective (near many different segments, where `near' is calibrated to the gripper radius -- $4.25$ cm or 20 pixels in our setup) and diverse (different candidate points are near different segments), ensuring that grasps can capture different segments.

%The method then uses the predictor for a fixed set of orientations (spaced equally) for each point and chooses the grasp that is predicted to capture the largest amount of segments.
% using the intuition that grasps can only capture garments whose distance is at most the gripper radius ($4.25\,$cm, corresponding to 20 pixels in our experiments). 
% An ideal set of candidate points has two features: the candidate points should be individually \emph{good}, and the set of candidate grasp points should be \emph{diverse}, as similar candidate points are generally redundant. A good grasp point is one which is near many different segments, where `near' is calibrated to the gripper radius, since a grasp at that point has the potential to simultaneously capture many garments; and a set of candidate points is diverse if different candidate points are near different segments.

\subsection{Depth-Based Methods}

Depth-based methods use the depth output of the RGBD overhead camera to select the next grasp. To solve the Teenager's Problem, these methods are used repetitively until the workspace is clear: at each step, we capture a new depth image of the scene, use one of the methods below to generate a new grasp, and then execute that grasp.
We examine two variations:

\subsubsection{Height} This variation selects the highest point in the scene and chooses the orientation to be that of the major axis of a local principal component analysis (PCA) around the grasp point.

\subsubsection{Volume} This variation considers the total volume of garments in a disc of radius $R$ around a candidate grasp point $(x,y)$, which is estimated by summing the heights of all the pixels within that radius, and then selects the point with the largest total volume. As in the Height method, orientation is selected using a local PCA.

%\subsubsection{Weight} This method selects the grasp point with the largest predicted garment weight, which is estimated via a learned Autoencoder model.

%\afterpage{\clearpage}
% \begin{figure}[H]
%    \centering
%     \includegraphics[width=\linewidth]{figures/consolidation_plan_and_actions_small.png}
%     \caption{An example execution using the consolidation method. The robot performs sequence of pick-and-place actions on the work surface to consolidate garments. It takes only five transport actions to clear all ten garments off the table.} 
%     \label{fig:result-consolidation} 
% \end{figure}

\subsection{Hybrid methods} \label{sec:hybrid}

One experimental observation was that depth-based methods and segment-based methods have different strengths---in particular, depth-based methods excel at picking occluded garments (which generally result in taller piles) while segment-based methods excel at simultaneously picking adjacent garments. This motivates considering \emph{hybrid} methods which make use of both depth data and segmentation data.

To take advantage of how these methods complement each other, hybrid methods do the following: given a height threshold (we use 0.1\,m), if the tallest pile is taller than the threshold, execute a (single) grasp as given by the depth-based method; if all piles are below the threshold, execute one cycle of the segment-based method.

\subsection{Segment-based method with consolidation}

Another avenue to improving OpT is to first consolidate the garments into large piles for transport to the basket; this can improve overall efficiency in cases where the basket is located at some distance from the work surface, making transports costly relative to manipulations within the workspace. An efficient primitive for consolidation is the \emph{grasp sequence} where each pick-and-place movement picks up where the last one placed; this both saves on robot movement time (no travel distance to the next pick point) and, ideally, allows the robot to accumulate more garments as it goes before depositing them in the basket.

An extension of the segment-based method above to include consolidations follows:
\begin{enumerate}
    \item generate the grasp plan as in the no-consolidation segment-based method (i.e., as described in Sec.~\ref{sec:segment-based}), and sort them in decreasing distance to the basket (i.e., the first grasp is the furthest one to the basket);
    \item for the next grasp in the plan, $i$, estimate the \emph{expected area} of grasped garments, using the formula $\sum_j p_{i,j} M_j$;
    \item if the \emph{expected area} for this next grasp will \emph{not} make the \emph{total expected area} exceed a pre-determined \emph{grasp area threshold}, then execute the next grasp, and add the \emph{expected area} to the \emph{total expected area}. If the expected area for the next grasp \emph{will} make the \emph{total expected area} exceed the threshold, then go back to step 2 above; 
    \item if no such grasp exists, transport the currently-held garments to the basket.
\end{enumerate}
Going from the furthest grasp point to the closest follows the intuition that a method using consolidation should consolidate towards the basket since this will always shorten the distance between the grasped garments and the basket, even if some are dropped along the way, and this will tend to compress them into a smaller space, facilitating later multi-object grasps.

\begin{figure}[H]
   \centering
    \includegraphics[width=\linewidth]{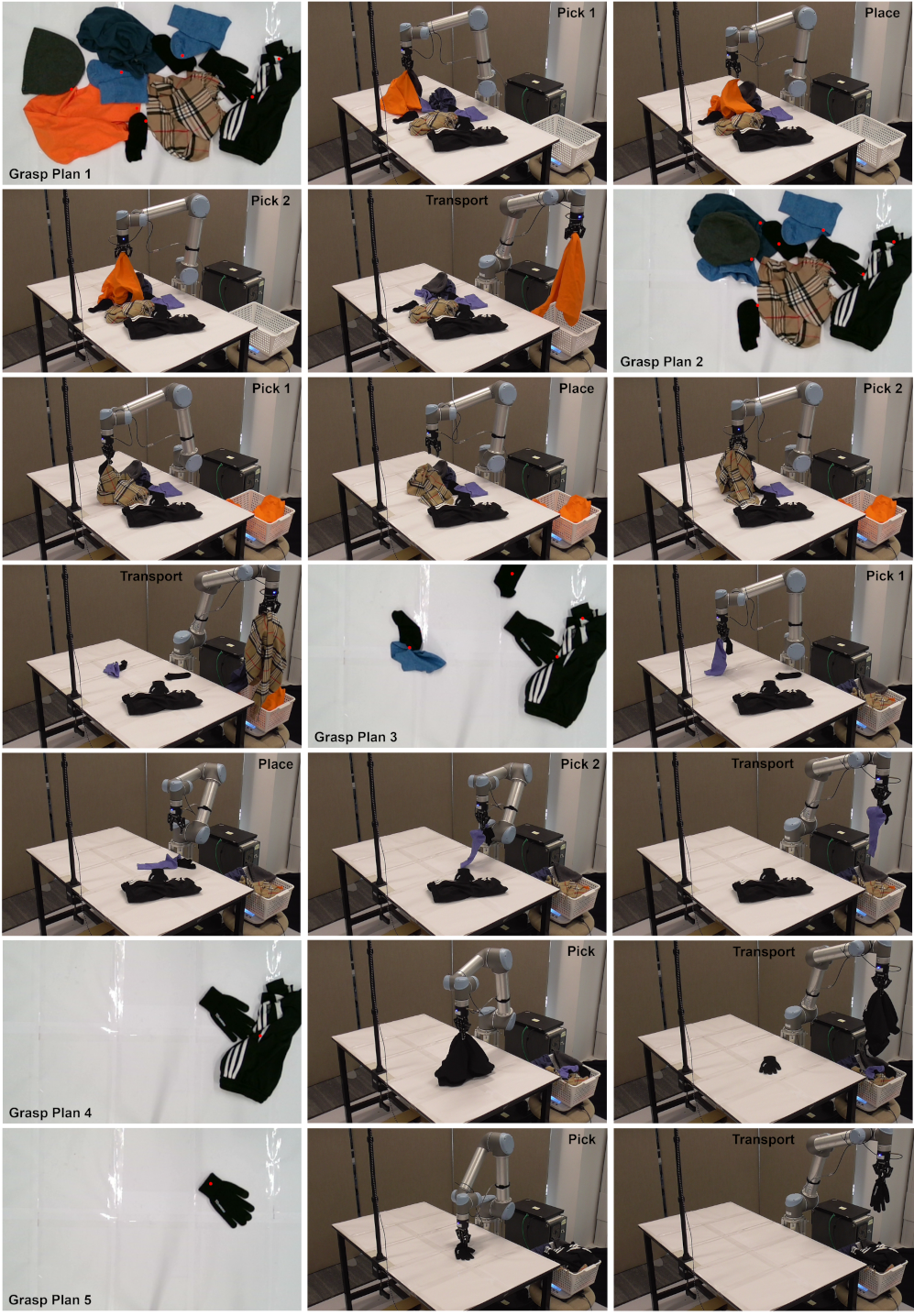}
    \caption{An example execution using the consolidation method. The robot performs sequence of pick-and-place actions on the work surface to consolidate garments. It takes only five transport actions to clear all ten garments off the table.} 
    \label{fig:result-consolidation} 
\end{figure}

The grasp area threshold corresponds to the intuition that the gripper has a limit to the amount of fabric it can hold and thus trying to accumulate more than that limit in one grasp is counterproductive. 

We show an example execution using the consolidation method in Fig.~\ref{fig:result-consolidation}. The trips to the basket are marked as ``Transport.'' The consolidation actions result in multiple garments being picked simultaneously, and significantly reduce the number of transport actions.

\subsection{Baseline} \label{sec:baseline}

Finally, as a baseline, we use the \emph{random} method, which uniformly randomly selects a garment point $(x,y) \in \cX_g$ with a uniformly random orientation $\theta \in [-\pi/2, \pi/2]$, accounting for the gripper's symmetry.

%Specifically, the method generates a Minkowski sum of each segment with a circular disk whose diameter is the gripper width in order to expand them. Thus, each point is considered `near' every segment whose expansion covers that point. This produces a function from points $(x,y) \in \cX$ to subsets of the masks

%% file: sections/05-NNs.tex
\begin{table*}[htb!] 
\centering
\caption{OpT (Objects per Transport) by grasping method}
%\vspace{0.5pc}
\begin{tabular}{c|cc|c|cc|c} 
    \toprule
   ~ & \multicolumn{2}{c|}{Depth-Based Methods} & ~ & \multicolumn{2}{c|}{Hybrid Methods} & ~ \\
     Random & Volume & Height & Segment & Volume & Height & Consolidation \\
     \midrule
    $1.07 \pm 0.07$ & $1.27 \pm 0.11$ & $1.22 \pm 0.07$ & $1.55 \pm 0.11$ & $1.34 \pm 0.10$ & $1.45 \pm 0.14$ & $\mathbf{1.94 \pm 0.15}$ \\
%        $1.48 \pm 0.11$ & $1.76 \pm 0.13$ & $1.77 \pm 0.20$ & $1.77 \pm 0.13$ & $1.87 \pm 0.13$ & $1.98 \pm 0.24$ & $2.47 \pm 0.2$ \\
    \bottomrule
\end{tabular} \label{tab:opt}
%\vspace{0.5pc}
%\caption*{hello}

\end{table*}

\section{Experiments} \label{sec:experiments}

\begin{figure}[b!]
    \centering
    % testgarms, cloth_set cloth_set_no_border
    % \includegraphics[width=\linewidth]{figures/cloth_set.png}
    \includegraphics[width=\linewidth]{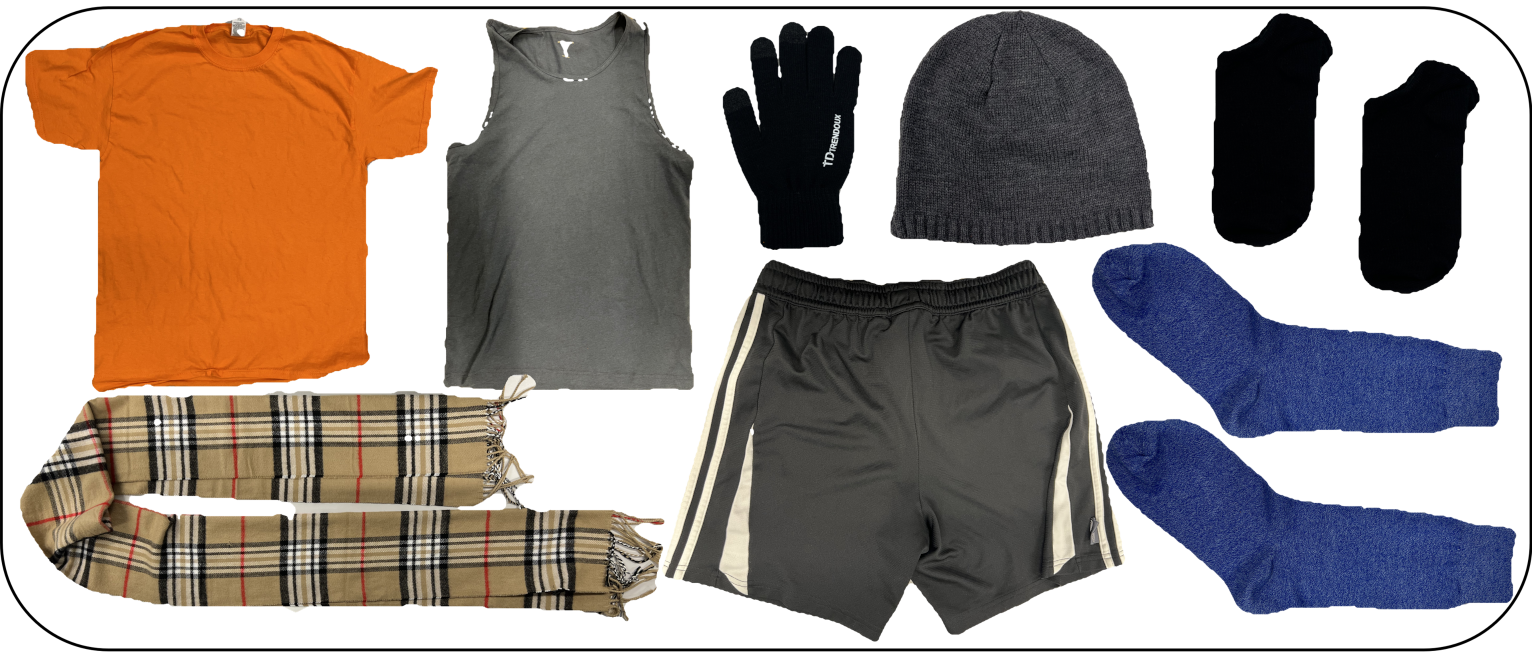}
    \caption{The test set of 10 garments, representing a variety of different sizes, weights, textures, colors, patterns, flexibility, and garment classes. Some garments have similar colors to present a challenge for segmentation. We also use long garments, such as the scarf, that present a challenge for grasping.} 
    \label{fig:test-set} 
\end{figure}

We tested all algorithms on the test set of 10 garments (see Fig.~\ref{fig:test-set}) with 25 sample runs. Each sample run begins with a randomized scene containing all 10 test set garments, and ends when the workspace is cleared of garments.

\subsection{Data collection pipeline} \label{sec:data-pipeline}

To run the experiments, we used a semi-autonomous data collection pipeline, in which experimental scene reset, randomization and data recording are done automatically, with the experimenter only needing to correct problems when they arise (for instance, if a garment falls off the work surface, the experimenter must return it for the next sample). The system uses the recorded weight data to automatically notify the experimenter when such a problem occurs, to minimize the amount of human attention necessary for data collection.

The scene is automatically reset in the following way:
\begin{enumerate}
    \item The robot grasps the basket and empties it over the work surface to deposit the garments, then places the basket back to its original position.
    \item The robot executes a sequence of random pick-and-place actions on the surface to shuffle the garments. Each pick position is randomly (uniform) sampled from the foreground cloth points $\cX_g$, and each place position is sampled from the full rectangular workspace (a rectangular region of 1 m $\times$ 0.6 m). In our experiments, 10 such moves were performed for every scene reset
\end{enumerate}
The experiment is then performed with the selected method, recording at each step the overhead RGBD output, the grasp location and orientation, and the weight of the garments in the basket. The experiment is paused for 3 seconds after every transport to allow the scale's output to settle. Full dataset is available at \url{https://sites.google.com/view/TeenSP2024}.

\subsection{Metrics evaluated}

Our metric for evaluating the methods is \emph{Objects per Transport} (OpT), which denotes the average number of objects taken during each transport and measures the general effectiveness of the performed grasps. We use OpT, as opposed to Picks Per Hour (PPH), as OpT directly measures grasp quality and PPH depends heavily on implementation details (particularly concerning computation time), which reduces its reliability as a metric in this setting. \revision{Additionally, maximizing OpT is a good strategy particularly when the target basket is far and consolidation is important.}

%(ii) \emph{Picks per Hour} (PPH), which measures the overall efficiency of each algorithm in the given setup. While increasing PPH is the ultimate goal of multi-object grasping, this work focuses on OpT as the primary metric for comparison, as PPH is greatly influenced by specific implementation details (which include computation time) and therefore not a reliable measure of the value of the proposed approaches.

For each algorithm tested, OpT was evaluated on all 20 sample runs, which were then averaged to yield the final result and $95\%$ confidence bounds.

% were estimated in the standard way, by taking $\sigma \frac{1.96}{\sqrt{n}}$, where $\sigma$ is the standard deviation of the metric over the sample runs and $n = 25$ is the number of sample runs.

% We use a pre-trained ResNet model to capture the RGB image features. 
% During the training phase, we utilize contrastive loss to enforce similarity among positive pairs (representing successful grasping actions) and dissimilarity among negative pairs (indicating ineffective attempts).
% To prepare input pairs for the Siamese network, we systematically generate masked images that correspond to distinct grasping actions, by employing a suite of data augmentation strategies, including translation, rotation, and cropping.

%% file: sections/06-Results.tex
\section{Results}
%[TODO: Build out results tables \& begin filling in]

The results of the experiments are given in \Cref{tab:opt},  
%As we investigated both the segmentation method and the depth methods by themselves and combined, we organize the tables to reflect this. 
%The method in the upper-left (`None' depth method, and `None' segment predictor) corresponds to the random baseline (\Cref{sec:baseline}), while entries that have both a segmentation and a depth component correspond to hybrid methods.
showing that both depth-based and segment-based methods yield clear improvements over the random baseline. In particular, the segment-based method provides additional $50\%$ OpT, achieving $1.57$ objects per transport to the basket. Although hybrid methods achieve better OpT compared to depth-based methods, they cannot beat the segment-based approach. %However, it is important to note that only the segment-based method achieves this without the use of depth information, which may not be available.
{\color{black} Finally, at the cost of both computational overhead and additional physical actions, the segmentation with consolidation method drastically improves OpT, achieving $1.91$ objects per transport, $81\%$ over the baseline.}

%It should be noted that while grasp quality is the focus and OpT is the most meaningful metric for this work, the ultimate goal remains improving pick efficiency as measured by PPH. The depth-based methods, which do not perform significant computations to find grasps, improve PPH from $477$ for the baseline to $526$ and $525$ for max-volume and max-height grasps respectively. While the segmentation method registers a slight decrease in PPH (to $453$), optimizing the methods' speed may increase the PPH up to a comparable $523$ (determined by subtracting computation time in the experiments) with no need for depth data.

Note that while grasp quality is the focus and OpT is the most meaningful metric for this work, another important metric for pick efficiency is PPH. The depth-based methods, which do not perform significant computations to find grasps, improve PPH from $125$ for the baseline to $147$ and $145$ for max-volume and max-height grasps, respectively. While the segment-based method requires more computation, it still registers a higher PPH of $174$. However, compared to the segment-based method, the consolidation approach registers a slight decrease in PPH (to $160$), even if it has a higher OpT score. This is due to the extra actions required by the consolidation method, and shows {\color{black} that its efficacy relative to the segment-based method depends on the basket being further from the workspace.} %again that its efficiency would be significant, only if the basket gets further to the workspace. 

%Finally, at the cost of both computational overhead and additional physical actions, the segmentation with consolidation method improves OpT drastically, achieving $1.91$ objects per transport, $81\%$ over the baseline.

%While PPH is not the primary metric as it does not directly measure grasp quality, in the experiments the 

%While increasing Picks Per Hour (PPH) is the ultimate objective, for these experiments it is too implementation-dependent to be a reliable measure (as it includes the computation time taken by unoptimized methods); OpT is preferred as it directly measures average grasp quality.

%For PPH with computation time, we find that the segment-based methods generally 

\subsection{Comparison of Methods}
What are the inherent advantages and disadvantages of the two approaches outlined above? 
\begin{itemize}
    \item Depth-based methods require both RGB and depth images to compute grasps. In contrast, segment-based methods only rely on RGB images, rendering them suitable for systems lacking depth cameras.
    \item Segment-based methods often require more computational resources, as they involve neural network-driven segmentation and subsequent cleanup. To ensure efficient computation without compromising speed, it may be necessary to deploy GPUs or opt for segmentation methods optimized for CPU processing.
     \item A notable challenge faced by segment-based methods is their limited ability to detect grasps that remove occluded garments. In contrast, depth-based methods more often grasp over occluded garments due to their utilization of depth information, which provides an enhanced perception of garment depth.
    \item Conversely, segment-based methods explicitly choose grasps to simultaneously capture garments situated closely together, whereas depth-based methods cannot determine which points are in proximity to multiple visible garments.
\end{itemize}
% This comparison suggests that while segmentation-based methods are generally more deployable due to not needing a depth camera, if one has access to depth data, further efficiency gains can be made by combining both depth and segmentation information in a single method.

% \begin{table}[htb]
% \centering
% \caption{Comparison Results of Objects per Transport (OpT) for Physical Decluttering Experiments}
% \begin{tabular}{cccccc}
% \begin{tabular}{p{1.4cm}p{0.9cm}p{0.9cm}p{0.9cm}p{0.9cm}p{0.9cm}}
%   \toprule
%     \multirow{2}{*}{} & \multirow{2}{*}{} & \multicolumn{4}{c}{Depth Method} \\
%     \cmidrule{3-6} & & None & Volume & Height & Weight \\
%   \midrule
%   % \multirowcell{3}{Segmentation\\Method}
%     \multirow{4}{*}{\shortstack{Segmentation\\Method}} & None & 1.431$\pm$0.0 & 1.699 & 1.653 & -\\
%     \cmidrule{3-6}                         & Baseline & 1.715 & 1.81 & 1.836 & -\\
%     \cmidrule{3-6}                         & NN & 1.618 & - & - & -\\
%   \bottomrule
% \end{tabular}
% \label{table:OpT}
% \end{table}

\subsection{How our system scales with the number of garments}
We performed additional experiments to measure how long it takes
(i) for the segment-based method to identify garments, and (ii) to solve the resulting MILP, as we increased the number of garments.
We present the results in Table~\ref{table:scaling}.
In the experiment, we incrementally added five items at a time to the scene, ensuring they remain non-overlapping to only measure the effect of number of garments on the performance. Still, the detected number of segments did not exactly match the number of garments, which are also shown in the table. While the segmentation time was not affected by the number of garments, the MILP solution time showed a faster than linear increase.  

We also performed an experiment where we maintained a constant number of garments (35) but increased the number of overlapping groups, where all objects within a group were in contact, to test how the system scaled with the degree of overlap of the garments in the scene. The results did not show an obvious relationship between the degree of overlap and segmentation or the MILP solution time.
\begingroup

\setlength{\tabcolsep}{10pt} % Default value: 6pt
\begin{table}[tb!]
\centering
\caption{Segmentation and MILP solution time with increasing number of garments}
\begin{tabular}{c c c c}
\hline
\# garments & \# segments & Seg. time (s) & MILP soln. time (s) \\ \hline
5             & 4             & 19.66       & 1.43           \\ 
10            & 9             & 18.86       & 3.88           \\ 
15            & 14            & 18.74       & 6.73           \\ 
20            & 19            & 18.73       & 9.95           \\ 
25            & 24            & 18.72       & 13.42          \\ 
30            & 29            & 18.81       & 16.86          \\ 
35            & 35            & 18.70       & 22.27          \\ \hline
\end{tabular}
\label{table:scaling}
\end{table}
\endgroup

%% file: sections/07-dirty-laundry-conclusion.tex
\section{Conclusion}
In this work, we tackle the challenging problem of robotic garment decluttering, by formalizing the Teenager's Problem and developing both depth- and segment-based methods to solve it.
%proposing innovative methods and learning models. 
%Our study encompassed both depth-based and segment-based approaches, aiming to minimize robot trips to the laundry basket and enhance the efficiency of grasping multiple garments simultaneously.
We use recent advances in image segmentation \cite{segment-anything} to explore an approach that uses it to distinguish garments in the image and find grasps that are likely to capture as many as possible.

%In this work [TODO]

\subsection{Dirty Laundry and Future Work}

However, this work has certain limitations and leaves a number of areas open for improvement:
\begin{itemize}
    % \item We assumed a flat work surface containing only garments that needed to be cleared away; however, a typical scene may include clothing strewn over furniture (such as chairs, beds and couches) as well as other objects whose inclusion in the laundry would be highly undesirable (e.g. wallets or phones). Thus, further work is needed to design a method that can work with garments placed on more complex surfaces and which can reliably distinguish between items that should and should not be placed in the laundry basket.
    \item All the proposed methods rely on accurately separating the garments (the foreground) from the table surface (the background) using the RGB image, which is done here via color thresholding. While this was reliable in our experimental setup, a different system may be needed if any garments are the same color as the background.
    % (though background subtraction is a known computer vision problem).
    % \item While the heap-based methods used the depth data and the garment-based methods made extensive use of the RGB images taken from the overhead camera, we did not consider methods which combined these.
    \item While all the methods considered here grasp at a fixed height above the work surface with a perfectly vertical gripper, 
    %(generally as close as possible without colliding with the surface, to allow grasping of garments lying flat on the surface), 
    %sometimes the most efficient grasp may be some distance above the work surface;
    the most efficient grasp may not share those characteristics. Additional improvements might be obtained by optimizing the grasp height or angle.%for instance, when the garment pile is too deep, grasping above the work surface may result in a collision with garments pressed against the work surface. While the Robotiq gripper used in this work will give way to prevent a forcible collision, it still loses some of its ability to grasp.
    %\item While segmentation-based methods have advantages, one disadvantage not apparent from the data is an increased sensitivity to camera calibration. Depth-based methods appear to be more robust to inaccuracies in camera calibration.
    \item While the $81\%$ OpT increase from the segmentation with the consolidation method is large, an average of roughly $2.1$ rearrangement actions were performed to consolidate prior to each transport saved over the baseline. This will increase efficiency in cases where transports are relatively costly, e.g. when the target basket is far from the workspace.
    %\item Following the above point, different variations of the garment decluttering method (for instance, switching between compression and consolidation steps in a different way or treating the heaps in a spatially determined order to minimize gripper travel time) were not considered, and the potential benefits of doing so were therefore not realized. 
\end{itemize}

%While this work provides a foundation for the study of multi-object grasping of deformable objects on a surface, many avenues of future work remain. First, the approaches studied in this work are not exhaustive, and more sophisticated methods, for instance by simulating the scene directly, may provide larger gains. Additionally, the Teenager's Problem is only one of many practical problems that can be addressed by multi-object grasping of deformable objects. 
In future work, we can explore extensions such as sorting of clothes (for instance, separating clothing by type or color). Additionally, although we present only the analytic grasp predictor, the segment-based method described in \Cref{sec:segment-based} is compatible with any grasp predictor, which could be improved using self-supervised data collection.

\section*{Acknowledgments}
M.~Dogar was supported by the UK Engineering and Physical Sciences Research Council [EP/V052659/1].